\pdfoutput=1

\documentclass[11pt]{article}

\usepackage[final]{acl}

\usepackage{times}
\usepackage{latexsym}

\usepackage[T1]{fontenc}

\usepackage[utf8]{inputenc}

\usepackage{microtype}

\usepackage{inconsolata}

\usepackage{graphicx}
\usepackage{booktabs}
\usepackage{appendix}
\usepackage{comment}
\usepackage{multicol}
\usepackage{multirow}
\usepackage{makecell}
\usepackage{float}

\newcommand{\tom}[1]{\textcolor{black}{#1}}
\newcommand{\am}[1]{\textcolor{black}{#1}}
\newcommand{\fr}[1]{\textcolor{black}{#1}}

\usepackage{enumitem}
\setlist[itemize]{leftmargin=*}
\title{Language is Scary when Over-Analyzed: Unpacking Implied Misogynistic Reasoning with Argumentation Theory-Driven Prompts}

\author{Arianna Muti$^1$,
  Federico Ruggeri$^2$,
  Khalid Al-Khatib$^3$,
  Alberto Barrón-Cedeño$^1$ \\
  \bf
 Tommaso Caselli$^3$ \\
  $^1$ DIT, University of Bologna \\
  $^2$ DISI, University of Bologna \\
  $^3$ University of Groningen \\
  $^1$\texttt{\{arianna.muti2, aikaterini.korre2, a.barron\}@unibo.it} \\
  $^2$\texttt{federico.ruggeri6@unibo.it} \\
  $^3$\texttt{\{khalid.alkhatib, t.caselli\}@rug.nl}
  }

\begin{document}
\maketitle
\begin{abstract}
We propose misogyny detection as an Argumentative Reasoning task and we investigate the capacity of large language models (LLMs) to understand the implicit reasoning used to convey misogyny in both Italian and English. The central aim is to \tom{generate} the missing reasoning link between a \tom{message and the implied meanings encoding the misogyny.} 
Our study uses argumentation theory as a foundation to form a collection of prompts in both zero-shot and few-shot settings. These prompts integrate different techniques, including chain-of-thought reasoning and augmented knowledge. Our findings show that LLMs fall short on reasoning capabilities about misogynistic comments and that they mostly rely on their implicit knowledge derived from internalized common stereotypes about women to generate implied assumptions, rather than on inductive reasoning. 
\end{abstract}

\section{Introduction}


According to the 7$^{th}$ Monitoring Round of the EU Code of Conduct on Countering Illegal Hate Speech Online,\footnote{\url{https://bit.ly/3yIRYWg}} Social Media are slowing down the removal of hateful content within 24 hours, dropping to 64\% from 81\% in 2021. The prevalence of hate speech phenomena has become a factor of polarization and pollution of the online sphere, creating hostile environments that perpetuate stereotypes and social injustice. 

Previous work on hate speech detection from the NLP community has contributed to definitions \cite{fortuna-etal-2020-toxic, pachinger-etal-2023-toward, korre-etal-2023-harmful}, datasets \cite{chiril-etal-2020-annotated,pamungkas_misogyny_2020,guest-etal-2021-expert,zeinert-etal-2021-annotating}, and systems \cite{caselli-etal-2021-hatebert,lees2022new}. However, most of these contributions have focused (more or less consciously) on explicit forms of hate. Recently, there has been an increasing interest in the study of implicit realization of hate speech phenomena \cite{caselli-etal-2020-feel,wiegand-etal-2021-implicitly-abusive,ocampo-etal-2023-depth}.

\begin{figure}[!t]
    \centering
    \includegraphics[width=1.05\linewidth]{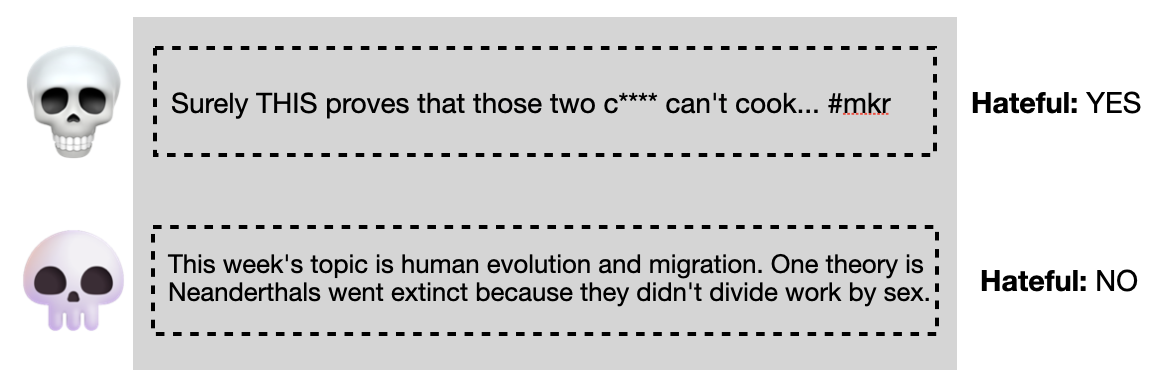}
    \caption{Results from \texttt{bert-hateXplain} model for explicit (\includegraphics[height=1.0em]{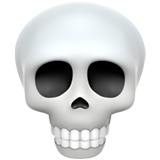}) \textit{vs}. implicit (\includegraphics[height=1.0em]{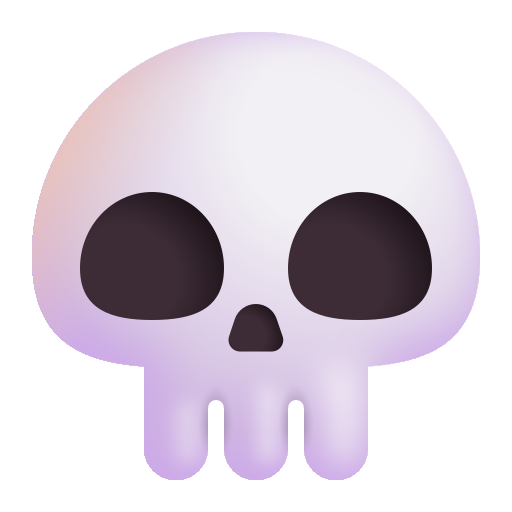}) misogynous messages. 
    }
    \label{fig:perspective_example}
\end{figure}

Implicit hate speech is more elusive, difficult to detect, and often hidden under apparently innocuous language or indirect references. 
These subtleties present a significant challenge for automatic detection because they rely on underlying assumptions that are not explicitly stated. As illustrated in Figure~\ref{fig:perspective_example}, the \texttt{bert-hateXplain} model\footnote{\url{https://huggingface.co/tum-nlp/bert-hateXplain}} correctly mark as hateful the explicit message (\includegraphics[height=1.0em]{img/skull_explicit.png}), but it fails with the implicit ones (\includegraphics[height=1.0em]{img/skull_implicit.png}).
To correctly spot 
the implicit message, the system would have to 
identify at least 
the implied assumptions that assume that ``\textit{women aren't as capable as men.}'' and \textit{``women should be told what to do''}.\footnote{Example and explanations extracted from~\citet{sap-etal-2020-social}.} 



In this contribution we investigate the abilities of large language models (LLMs) to correctly identify implicit hateful messages expressing misogyny in English and Italian. 
In particular, we explore how prompts informed by Toulmin's Argumentation Theory~\cite{toulmin1979introduction}
are effective in reconstructing the \textit{warrant} needed to make the content of the messages explicit and thus facilitate their identification as hateful messages~\cite{kim-etal-2022-generalizable}. 
By prompting LLMs to generate such warrants, we further investigate whether the generated texts are comparable to those of human annotators, thus offering a fast and reliable solution to enrich hateful datasets with explanations and contributing to improve the generalization abilities of trained tools. 
We summarize our contributions as follows:

\begin{itemize}  \setlength\itemsep{0em}
    \item We present a novel formulation of implicit misogyny detection as an Argumentative Reasoning task, centered on reconstructing implicit assumptions in misogynous texts (\S\ref{sec:problem-formulation}).
    \item We introduce the first dataset for implicit misogyny detection in Italian (\S\ref{sec:datasets}).\footnote{All data will be released via a Data Sharing Agreement.} 
    \item We carry out an extensive set of experiments with two state-of-the-art instruction-tuned LLMs (\texttt{Llama3-8B} and \texttt{Mistral-7B-v02}) on English and Italian datasets (\S\ref{sec:experiments}).
    \item We conduct an in-depth qualitative analysis of the automatically generated implicit assumptions 
    against 300 human-generated ones (\S\ref{sec:results}).
    
\end{itemize}

\section{Related Work}
\label{sec:rel_work}


Hate speech detection is a widely studied research area, covering different targets and linguistic aspects.
We discuss literature work on implicit misogyny detection with particular attention to contributions in reconstructing implicit content.

\paragraph{Implicit Hate Speech Detection}

Hate Speech Detection is a 
popular research domain with more than 60 datasets covering distinct targets (e.g., women, LGBTIQ+ people, migrants) 
and forms of hate (e.g., sexism, racism, misogyny, homophobia)
in 25 languages, according to the Hate Speech Dataset Catalogue.\footnote{\url{https://hatespeechdata.com}}
In its early stages, but still predominant nowadays, research in this domain focused on developing datasets for detecting explicit cues of hate speech, like messages containing slurs or swear words~\cite{jahan-oussalah-2023-hate-speech}.
However, hate speech is often implicit, characterized by the presence of code language phenomena such as sarcasm, irony, metaphors, circumlocutions, and obfuscated terms, among others~\cite{waseem-etal-2017-understanding, wiegand-etal-2021-implicitly-abusive}.
For this reason, implicit hate speech detection has progressively gained momentum in recent years, and several efforts have been put into the development of datasets for this purpose~\cite{sap-etal-2020-social,elsherief-etal-2021-latent,hartvigsen-etal-2022-toxigen,ocampo-etal-2023-depth}.
A relevant feature of these datasets is the presence of implied statements in free-text format, which contributes to explaining the content of hate speech messages.
While the use of these statements has been shown to have a positive effect on classification performance~\cite{kim-etal-2022-generalizable,kim-etal-2023-conprompt}, few efforts have been put in automatically generating such implied assumptions~\cite{elsherief-etal-2021-latent}. 
As \citet{yang-etal-2023-hare} point out, current annotation schemes in this area present significant reasoning gaps between the claim and its implied meaning. 
Moreover, no effort has been made to evaluate widely adopted LLMs on their reasoning capabilities required to generate high-quality implied assumptions.
To the best of our knowledge, our work is the first study to propose an empirical evaluation of LLMs for implicit misogyny detection and the generation of explanations for Italian and English. 
Available datasets targeting misogyny in Italian~\citep{fersini2018AMI,fersini2020ami} are highly biased toward explicit messages, with very few messages that qualify as implicit. 
To fill this gap, we have developed the first Italian dataset for this task, ImplicIT-Mis.
In our work, we define misogyny as a property of social environments where women perceived as violating patriarchal norms are ``kept down'' through hostile or benevolent reactions coming from men, other women, and social structures~\citep{Lopes2019-LOPPTP,1f4bc88c18994d6985d05fa789ee274e}, going beyond the simplistic definition of misogyny as hate against women.


\paragraph{Implied Assumptions Generation}

The implied assumptions instantiate statements that are presupposed by the implicit hate speech message. This can be seen as the elicitation of implicit knowledge, corresponding to new content semantically implied by the original message~\cite{srikanth-li-2021-elaborative,zaninello-magnini-2023-smashed}. 
Although limited, previous work on the generation of implied meanings ---usually in the forms of explanations--- has moved away from template-based methods~\cite{10.1145/2600428.2609579} to the application of encoder-decoder or decoder-only models~\cite{saha-etal-2021-explagraphs,xing-etal-2022-automatic,cai-etal-2022-multi}. Generating explanations for implicit content poses    
multiple challenges concerning the quality of the generated texts, whose primary goal is to be reasonable and informative. 
Some approaches generate explanations by identifying pivotal concepts in texts and linking them through knowledge graphs~\cite{ji-etal-2020-generating}
More recently, the underlying concepts are generated by directly querying LLMs~\cite{10.5555/3495724.3497422,fang-zhang-2022-data,yang-etal-2023-hare}.  
In this work, we follow the idea of using LLMs to identify the implied assumptions in the implicit messages, but rather than centering the reasoning process on identifying specific concepts, 
we formulate the problem as an Argumentative Reasoning task and apply Toulmin's Argumentation Theory~\cite{toulmin-1958-argument}.

\section{Misogyny Detection as Argumentative Reasoning Understanding} \label{sec:problem-formulation}


The elusiveness of implicit hate speech is due to its ambiguity. 
Implicit messages could be understood as critiques, opinions, or statements (see Figure~\ref{fig:perspective_example}) rather than as hateful. 
Hate, in this case, is expressed by assuming social biases, stereotypes, and prejudices against a specific target, women in the case of misogyny.
The identification of these assumptions requires access to the reasoning process behind arguments and opinions.  

Argumentative Reasoning (AR) offers a solution. 
AR
relies on the notion of an argumentative model or scheme, i.e. a formal representation of arguments into intrinsic components and their underlying relations. It aims at explicating an argument through the identification of its constituent components and relations~\cite{lawrence-reed-2019-argument}. For instance, the Toulmin's AR model organizes arguments into fundamental elements such as claim, warrant, and reason. AR models have been successfully applied in many NLP tasks, from Argument Mining~\cite{stab-gurevych-2017-parsing,habernal-gurevych-2017-argumentation,lauscher-etal-2018-argument} to warrant and enthymeme reconstruction~\cite{reisert-etal-2015-computational,boltuzic-snajder-2016-fill,habernal-etal-2018-argument,tian-etal-2018-ecnu,chakrabarty-etal-2021-implicit,bongard-etal-2022-legal}, argumentative scheme inference~\cite{feng-hirst-2011-classifying}, and fallacy recognition~\cite{habernal-etal-2018-name,delobelle-etal-2019-computational,goffredo-etal-2022-fallacious-argument,mancini-etal-2024-multimodal}.
%

\begin{figure}[!t]
    \centering
    \includegraphics[width=0.95\linewidth]{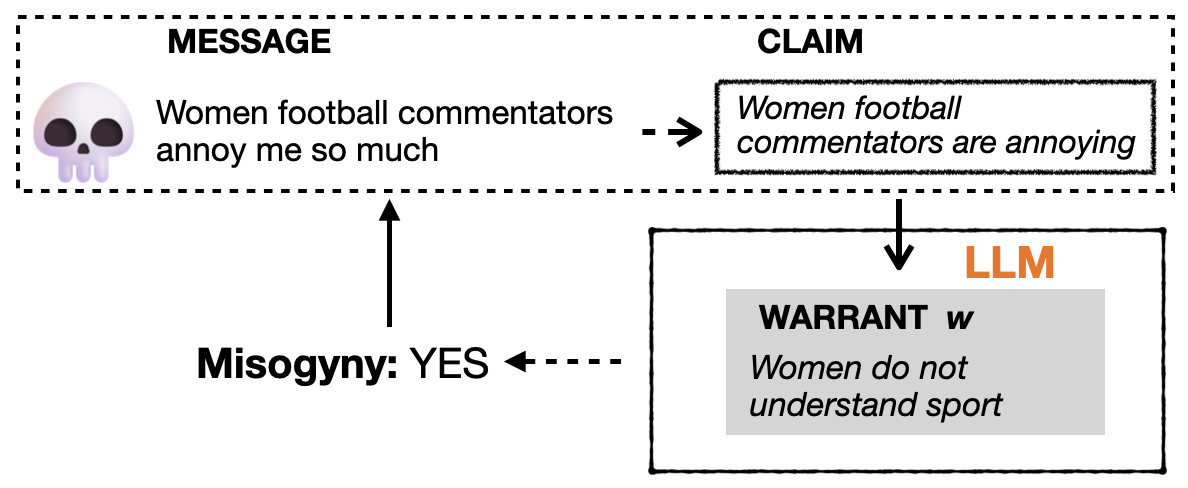}
    \caption{Example of a warrant (implicit logical connection) for an implicit misogynous message. Example and warrant are extracted from SBIC~\cite{sap-etal-2020-social}.
    }
    \label{fig:diagram}
\end{figure}

Grounded on previous work on AR in user-generated content~\cite{boltuzic-snajder-2016-fill,becker-etal-2020-implicit}, we frame implicit misogyny detection as an AR task~\cite{habernal-etal-2018-argument} based on the Toulmin's theory~\cite{toulmin1979introduction}, with the aim of developing more robust detection tools by explicitly describing the underlying reasoning process in these messages. More formally, let $c$ be the claim associated to a given message 
and $W$ = \{$w$$_1$, \dots $w$$_n$ \} be a set of possible warrants, i.e., logical statement(s) that support $c$.
The model must generate an associated 
\textit{w} and, based upon it,  
provide 
the requested classification
: whether the message is 
misogynous or not.

Figure~\ref{fig:diagram} graphically represents the approach described above. 
In this particular case, the generalization that women do not understand sport because it is stereotypically for men 
is what distinguishes a personal attack from a case of misogyny. 

While there have been efforts on evaluating LLMs in argumentative tasks, such as quality assessment~\cite{wachsmuth-etal-2024-argument-quality}, component detection~\cite{chen-etal-2023-exploring-potential}, and argumentative linking~\cite{gorur-etal-2024-relation-based}, the capability of LLMs for implicit argumentative reasoning has yet to be explored. To the best of our knowledge, our work is the first to assess LLMs on implicit misogyny through the lens of AR.

\section{Data}
\label{sec:datasets}


This section introduces the datasets used in our experiments. For Italian, the newly created ImplicIT-Mis corpus 
(\S~\ref{sec:implicit}). For English, SBIC+, an extended version of the \textsc{social bias inference corpus}~\cite{sap-etal-2020-social} enriched with misogynous texts from \textsc{Implicit Hate Corpus}~\cite{elsherief-etal-2021-latent} (\S~\ref{sec:bias_frames}).


\subsection{The ImplicIT-Mis Corpus}
\label{sec:implicit}

ImplicIT-Mis is a new manually collected and curated dataset for implicit misogyny detection in Italian. It consists of 1,120 Facebook comments as direct replies to either women-related news articles or posts on public pages of communities known to tolerate misogyny. 
An in-domain expert, who has been the target of misogyny, conducted the manual collection.
\footnote{The annotator is also an author of this paper.} This is in line with a participatory approach to NLP where the communities primarily harmed by specific forms of content are included in the development of datasets addressing these phenomena~\cite{caselli-etal-2021-guiding,abercrombie-etal-2023-resources}. For each comment, we keep 
 source (either a newspaper or a Facebook page) and its context of occurrence (the news article or the main post). 
All instances in ImplicIT-Mis are misogynistic.

The collection period ran from November 2023 to January 2024. 
We selected 15 Facebook pages of news outlets covering the whole political spectrum as well as different levels of public outreach (national \textit{vs.} local audiences), and 8 community pages. 
ImplicIT-Mis is organized around 104 source posts; 70\% of the 1,120 messages are comments to news articles from two national newspapers (\textit{La Repubblica} and \textit{il Messagero}). The full overview 
is 
in Appendix~\ref{sec:app_it_data}. 
On average, each comment is 19 tokens long, with the longest having 392 tokens and the shortest only one. An exploration of the top-20 keywords, based on TF-IDF, indicates 
a lack of
slurs or taboo words, confirming the quality of our corpus for implicit misogyny.

ImpliciIT-Mis is enriched with one annotation layer targeting the implied assumptions, as defined in $\S$\ref{sec:rel_work}. A subset of 150 messages was annotated by three Italian native speakers who are master students in NLP. Each annotator has worked on 50 different messages. On average, annotators took 2 hours to complete the task. The annotation guidelines for the generation of the implied assumptions are in Appendix~\ref{sec:guidelines_it}. 
We evaluated the annotators' implied assumptions against those of an expert (a Master student in gender studies and criminology). We used a subset of 75 sentences (25 from each annotator) and computed two metrics: 
BLEU~\cite{papineni-etal-2002-bleu} 
and BERTScore~\cite{bert-score}. 
These measures offer insight into how similar the human written implied assumptions are. 
We have obtained a BLEU score of 0.437 and an F1-BERTScore of 0.685 by combining all annotations. 
As the scores indicated, our pool of annotators tends to write the implied assumption adopting different surface forms, 
but with a similar semantic content, as suggested by the F1-BERTScore. 
Although implied assumptions have to be inferred, and therefore, humans need to interpret the text, they tend to come to the same conclusions. 
In the final version of the data, all manually generated implied assumptions have been retained as valid, meaning that for 150 messages, we have a total of 225 implied assumptions.



\subsection{\textsc{SBIC+}}
\label{sec:bias_frames}

SBIC+ is a dataset of 2,409 messages for implicit misogyny in English obtained by merging together 2,344 messages from SBIC and 65 from the \textsc{Implicit Hate Corpus} (IHC).

The \textsc{social bias inference corpus} (SBIC)~\cite{sap-etal-2020-social} 
consists of
150$k$ structured annotations of social media posts for exploring the subtle ways in which language can reflect and perpetuate social biases and stereotypes. It covers over 34$k$ implications about a thousand demographic groups. SBIC is primarily composed of social media posts collected from platforms like Reddit and Gab, as well as websites known for hosting extreme views, such as Stormfront. 

The structured annotation approach implies that different annotation layers are available to annotators according to their answers. The annotation scheme is based on social science literature on pragmatics and politeness. We retain all messages whose annotation for the target group was ``women'' or ``feminists'' and were labelled as hateful. We further cleaned the data from instances labeled as targeting women but were actually targeting other categories, like gay males. We also filtered out all texts containing explicit identity-related slurs to keep only implicit instances. For each message, we also retained all associated ``target stereotype'' which correspond to the warrants. 

The \textsc{Implicit Hate Corpus} (IHC)~\cite{elsherief-etal-2021-latent} contains 6.4$k$ implicitly hateful tweets, annotated for the target (e.g., race, religion, gender). The corpus comprises messages extracted from online hate groups and their followers on Twitter. 
Tweets were first annotated through crowdsourcing into explicit hate, implicit hate, or not hate.  Subsequently, two rounds of expert annotators enriched all implicit messages with categories from a newly developed taxonomy of hate, for the target demographic group, and the associated implied statement (i.e., the warrant in our framework). 
We have selected only tweets whose target demographic group was ``women''.


\section{Experimental Setup}
\label{sec:experiments}

\fr{Our main goal is to evaluate the abilities of models to generate the implied assumptions for implicit misogynous messages. 
By doing so, we can also evaluate the implicit knowledge of LLMs, for instance, named entities or events mentioned in texts. 
If they are not known, it would be impossible to understand the misogynistic nature of such texts. }
Each batch of experiments aims to address two tasks: (i) the generation of the implied assumptions or warrants and (ii) the classification of the messages as misogynous or not.
Regarding (i), we experiment with two prompting strategies: instructing the model to reconstruct the implied assumptions (\textbf{Assumption}) and the implicit claim $c$ and related warrants $W$ (\textbf{Toulmin}). 
We address these tasks both in a zero-shot and in a few-shot setting. 
While implied assumptions are generally broader than warrants, warrants specifically bridge the reasoning gap between claims and evidence. In our prompts, implied assumptions and warrants appear quite similar. Nevertheless, the use of these terminologies may significantly impact the model's behavior due to its sensitivity to prompt phrasing, therefore we experiment with both.

We experiment with two state-of-the-art LLMs: \texttt{Llama3-8B} and \texttt{Mistral-7B-v02}.%
\footnote{Refer to \url{https://huggingface.co/meta-llama/Meta-Llama-3-8B} and \url{https://huggingface.co/mistralai/Mistral-7B-Instruct-v0.2}}
For both, we select their instruction-tuned version. 
During preliminary experiments with 50 instances, we also tested Italian-specific LLMs, namely \texttt{LlaMantino}, \texttt{Fauno}, and \texttt{Camoscio}. They were all unable to generate valid implied assumptions, so we discarded them.
We consider the following baselines: (i) finetuned encoder-based models; (ii) zero-shot classification with LLMs; and (iii) few-shot classification with LLMs without generating explanations.


\paragraph{\texttt{Llama3-8B}}
The Llama3 series has several improvements over preceding versions, including a better tokenizer with a vocabulary of 128$k$ tokens, extended training on 15$T$ tokens, and grouped query attention for efficiency.
Around 5\% of the pre-training data concerns more than 30 languages, including Italian.
All Llama3 models have undergone safety fine-tuning for safeguarding the generation process over harmful content. This could trigger instances of over-safety, with the model being unable to follow the instructions and thus failing to provide a valid answer for our task.

\paragraph{\texttt{Mistral-7B-v02}} 
A competitive fine-tuned version of Llama2 using group-query attention, developed by MistralAI~\cite{Mistral:2023}.
In particular, the 7B version has been reported to obtain better performances when compared to \texttt{Llama2-7B} and \texttt{Llama2-13B}.
While details about the fine-tuning data are lacking, in our experiments, we observe that the model is responsive to Italian prompts.
The instruct-based versions of the models do not present any moderation mechanism. We thus expect this model to avoid over-safety and always return an implied statement and a classification value.


\subsection{Prompting Techniques}

Among recent prompting techniques, we selected \textbf{Chain-of-Thought} (CoT) and \textbf{Knowledge Augmentation}. CoT was chosen for its notable success in reasoning tasks \cite{lyu-etal-2023-faithful}. 
On the other hand, Knowledge Augmentation has been observed to reduce hallucinations and enhance contextual depth in model prompts, facilitating the generation of sophisticated outputs beneficial for tasks requiring substantial domain knowledge and nuanced reasoning \cite{minki:2023}.
Both techniques align with our goal of generating implicit components of arguments (implicit warrants) and support the construction of encoded warrant blocks. 
To the best of our knowledge, these techniques have not been used yet for a computational argumentation task, which makes them worth investigating. The full list of prompts can be found in Appendix \ref{sec:prompts_en} and \ref{sec:prompts_it}.
More in detail, CoT sequentially guides the model through a series of reasoning steps before arriving at a final answer or conclusion~\cite{wei:2024}. 
By following this structured approach, CoT prompts allow the identification of how the model's reasoning process influences its conclusions. This capability is particularly useful for reconstructing warrants that underlie the model's interpretations in our specific task. 

Knowledge-augmented prompting generates knowledge from an LLM and incorporates it as additional input for a task~\cite{liu-etal-2022-generated}. In our task, the generated knowledge serves as either the implied assumption or the warrant that we inject into the prompt to inform the classification.


\section{Results}
\label{sec:results}

We report two blocks of results: the first block focuses on \textbf{classification} of the messages. Since both the Italian and the English datasets contain only positive classes, we only report the Recall. The classification task offers an indirect evaluation on the goodness of the AR methods. The second block targets the \textbf{generation} of the implied assumptions/warrants. Considering the complexity and the pending issues related to the evaluation of automatically generated text~\cite{chang-etal-2024-survey-evaluation}, we report the results using established automatic metrics (i.e. BERTScore and BLEU) as well as a manual validation on a subset of 300 messages (150 per language) (\S~\ref{sec:results:qualitative-analysis}). The overall evaluation procedure we have devised allows us to assess both the performance of the models' in detecting implicit misogyny and the alignment between LLMs and human annotators in generating reasoning-based explanations.

All answers from LLMs have undergone post-processing to evaluate them properly. Two main post-processing heuristics concern the treatment of the ``refusal to provide an answer'' (including the refusal to generate the warrants) and the ``need of more context''. We considered both cases as if the messages were marked as not misogynous. While \texttt{Llama3-8B} tends to return refusals to answers, mostly due to the safeguard layer, \texttt{Mistral-7B-v02}  has a tendency towards indecisive answers requiring more context. \texttt{Llama3-8B} always provides an answer when applied to the Italian data. 
For completeness, Appendix~\ref{tab:appendix:results_overview} includes the results considering these cases as correct.
%

\begin{table}[!t]
    \centering
    \small
    \begin{tabular}{l@{\hspace{1mm}}l@{\hspace{0mm}}c@{\hspace{2mm}}c}
    \toprule
     \bf Setting & \bf Model & \bf ImplicIT-Mis & \bf SBIC+ \\ \midrule
    \multirow{2}{*}{fine-tuning} & \texttt{bert-hateXplain} & -- & 0.342 \\
    & \texttt{ALBERTo} & 0.380 & -- \\ \midrule \midrule
    \multirow{2}{*}{zero-shot} & \texttt{Llama3-8B} & 0.588 & 0.609 \\
    & \texttt{Mistral-7B-v02} & 0.050 & 0.319 \\ \midrule
    \multirow{2}{*}{few-shot} & \texttt{Llama3-8B} & \bf 0.738 & \bf 0.719 \\ 
    & \texttt{Mistral-7B-v02} & 0.259 & 0.416 \\ \midrule \midrule

    \multirow{2}{*}{\makecell[l]{zero-shot \\ Assumption}} & \texttt{Llama3-8B} & 0.542 & 0.448 \\
    & \texttt{Mistral-7B-v02} & 0.050 & 0.259  \\ \midrule
    \multirow{2}{*}{\makecell[l]{few-shot \\ Assumption}} & \texttt{Llama3-8B} & 0.480  & 0.616 \\
    & \texttt{Mistral-7B-v02} & 0.461 & \underline{0.685} \\ \midrule \midrule

    \multirow{2}{*}{\makecell[l]{zero-shot \\ Toulmin}} & \texttt{Llama3-8B} & 0.557 & 0.452 \\
    & \texttt{Mistral-7B-v02} & 0.346 & 0.374 \\ \midrule
    \multirow{2}{*}{\makecell[l]{few-shot \\ Toulmin}} & \texttt{Llama3-8B} & \underline{0.725} & 0.594 \\
    & \texttt{Mistral-7B-v02} & 0.556 & 0.604 \\ 
    \bottomrule
    \end{tabular}
    \caption{Classification results on ImplicIT and SBIC+. Best results in bold; second best underlined.
    }
    \label{tab:results_overview}
\end{table}

\subsection{Classification Results}
\label{sec:classification_results}

Table~\ref{tab:results_overview} summarizes the results for the classification task. With few exceptions - mostly related to \texttt{Mistral-7B-v02} - LLMs generally perform better than finetuned models. All few-shot experiments outperform their zero-shot counterpart, and \texttt{Llama3-8B} consistently performs better than \texttt{Mistral-7B-v02}. The best results are obtained by \texttt{Llama3-8B} with few-shot and no generation of either the implied statements or the warrants. However, for Italian, the \texttt{Llama3-8B} with the Toulmin warrant in few-shot achieves very competitive results (R=0.725). For English, on the other hand, the results are affected by the post-processing heuristics. Had we considered as correct the ``refusal to answer cases'',  the best score for English would have resulted in \texttt{Llama3-8B} few-shot with implied assumption (R=0.913).

In all zero-shot settings, the prompt based on Toulmin's warrant outperforms the prompt based on implied assumptions. In the few-shot settings, in ImplicIT-Mis, we observe a dramatic increase when switching from implied assumptions to Toulmin's warrant, with a performance gain of 24 points. On the contrary, on English, the warrant-based prompt falls behind.

\subsection{Implied Assumptions and Warrants Generation} \label{sec:results:qualitative-analysis}

\begin{table}[!t]
    \centering
    \small

\begin{tabular}{l@{\hspace{1mm}}l@{\hspace{2mm}}cccc}
\toprule
\bf Setting & \bf Model & \multicolumn{2}{c}{\bf BERTScore} & \multicolumn{2}{c}{\bf BLEU} \\ 
            &                 &\bf EN    & \bf IT   & \bf EN    & \bf IT \\\midrule
\multicolumn{2}{l}{\bf Assumption} \\
zero-shot   & \texttt{Llama3-8B} & 0.820 & - & 0.201 & -\\
few-shot    & \texttt{Llama3-8B}  & \bf 0.830 & - & 0.744 & -\\
            & \texttt{Mistral-7B-v02} & 0.823 & \bf 0.601& 0.361  & 0.240\\
\multicolumn{2}{l}{\bf Toulmin} \\
zero-shot   & \texttt{Llama3-8B} & 0.817 & 0.570 & 0.543 & 0.104\\
            & \texttt{Mistral-7B-v02} & 0.812 & 0.579 & 0.303 & 0.077\\
few-shot    & \texttt{Llama3-8B}      & 0.817 & 0.570 & \bf 0.871 & 0.261\\
            & \texttt{Mistral-7B-v02} & 0.813 & \bf 0.601 & 0.396 & \bf 0.313 \\

\bottomrule
\end{tabular}
    \caption{Automatic evaluation metrics for the best models generating implied assumptions/warrants (selection based on classification results).}
    \label{tab:metrics_best_only}
\end{table}

Table~\ref{tab:metrics_best_only} gives an overview of the evaluation using BERTScore and BLEU for the best models for English and Italian. While for SBIC+ every message has an associated explanation, for ImplicIT-Mis, only 150 messages present the implied assumptions. 
When \texttt{Llama3-8B} is asked to elaborate on the implied assumption in both zero- and few-shot settings, it does not follow the instruction, and only in 87 and 71 instances for Italian and English, respectively, generates a response.
In all the other cases, the model just answers the final question of whether it is misogynistic; therefore, we exclude them from the evaluation.
We also exclude all the results that do not reach at least a recall of 0.3 due to their low quality, as confirmed by manual inspection. 
\am{All BERTScores in English are around 0.81-0.83, showing high similar content between the human-written texts and the answers generated by the models. Therefore, both the implied assumptions and the warrants are aligned with those written by humans. In Italian the scores drop to 0.57-0.60. In terms of BLEU scores, the highest scores for English are produced by \texttt{Llama3-8B} few-shots with warrants, which shows an alignment with humans in terms of word choices. For Italian the scores are much lower, probably because of many wrong translations and lack of Italian references which cause wrong inferences.}

\subsection{Manual Validation} 

We further validate the generated implied assumptions and warrants by manually exploring a subset of 300 messages, 150 per language. 
For ImplicIT-Mis, we use the manually annotated instances, while we randomly extract 150 instances for SBIC+. 
We focus only on the best models: \texttt{Llama3-8B} few-shots warrant for ImplicIT-Mis and \texttt{Mistral-7B-v02} few-shots implied assumptions for SBIC+.

Overall, we find that 35\% of the generated warrants for ImplicIT-Mis are correct and 32\% lead to a correct classification of the messages. For SBIC+, the percentage of valid implied assumptions leading to a correct classification is 50\%, while correct implied assumptions leading to a wrong classification are 52\%. 
However, in  Italian all the correctly predicted examples were actually predicted for the wrong reasons, while in English this happened 37\% of the time. 
Therefore, we conclude that a correct explanation does not necessarily lead to a correct classification of misogyny, and this is always the case in the subsample we manually evaluated for Italian. \am{This can be seen as an evidence that the model relies on their internalized knowledge and spurious correlations to address the task and shows no reasoning skills, since the Italian texts, being collected to address this task, requires much more reasoning to be understood.}

We design a taxonomy to regroup all errors for both models.
We identify seven kinds of common errors in warrant and implied assumption generation. Table~\ref{tab:examples} provides some examples. 
Notice that, although all error categories lead to wrong implied assumptions/warrants, we decide to keep a general ``wrong inference'' as a valid category for all the cases that do not fall under any other category or there is no evident reason.

\begin{table}[t]

\begingroup
\renewcommand{\arraystretch}{1.5} 

\centering
\footnotesize
\resizebox{1.0\linewidth}{!}{%
\begin{tabular}{p{1.1\columnwidth}}
\toprule
\bf Sarcasm/Irony \hfill $\rightarrow$ IT: 2; EN: 26   \\
\textbf{Example:} \textit{my wife gets angry when i call her pet names apparently sandwich making machine is sexist...} \\
\textbf{Generated:} \textit{The text mentions a personal interaction between the speaker and their wife, and does not make a generalization about women} \\
\hline 

\bf Metaphorical and Figurative Language\hfill $\rightarrow$ IT: 14; EN: 1   \\
\textbf{Example:} \textit{Put a stuffed animal between them and they fight over it} \\
\textbf{Generated:} \textit{Men argue easily} \\
\hline

\bf Wrong Translations \hfill $\rightarrow$ IT: 10; EN: N/A  \\
\textbf{Example:} \textit{la lagna di mezzanotte} [The midnight complain] \\
\textbf{Generated:} \textit{There is a belief in witches and the existence of a ``witches' hour''. [...] The use of the term ``witches'' is likely a poetic or metaphorical way of describing challenges.} \\
\hline

\bf Opposite Intention \hfill $\rightarrow$ IT: 12; EN: 9 \\
\textbf{Example:} \textit{good girls don't get raped.}  \\
\textbf{Generated:} \textit{Women have control over whether or not they get raped.}\\
\hline

\bf Denial of Misogyny \hfill $\rightarrow$ IT: 3; EN: 4 \\
\textbf{Example:} \textit{doesn't society's acceptance of makeup for women give them a gender imbalance and greater attractiveness in aggregate over men, therefore inequality?}  \\
\textbf{Generated:} \textit{ Women are judged and valued primarily based on their physical appearance. } \\
\hline

\bf Lack of Reference \hfill $\rightarrow$ IT: 17; EN: 0 \\
\textbf{Example:} \textit{Pozzi era meglio, non cantava ma\dots il lavoro era lo stesso.} [Pozzi was better, she didn't sing but ... the job was the same]  \\
\textbf{Generated:} \textit{Pozzi is a better singer than the other one} \\
\hline

\bf Wrong Inference \hfill $\rightarrow$ IT: 42; EN: 34    \\
\textbf{Example:} \textit{i can't decide if today is a good or bad day my ex got hit by a bus, but i lost my job as a bus driver
}  \\
\textbf{Generated:} \textit{The text does not provide any assumptions about women or gender. The text is about the speaker's personal experiences and emotions.} \\
\bottomrule
\end{tabular}
}
\endgroup

\caption{Error categories in warrant generation. For each category, we report an input example, the corresponding LLM generation, and the category's distribution in Italian and English evaluation samples.}
\label{tab:examples}
\end{table}

\paragraph{Sarcasm/Irony} 
This is a common error in English, due to the relatively high number of jokes in SBIC+.
In these cases, the LLMs fail to capture the sarcastic/ironic intended meaning of the message and go for a more literal interpretation. 

\paragraph{Metaphorical and Figurative Language.}
This category indicates a failure to interpret another level of non-literal meaning. 
We have observed a much more frequent occurrence in Italian - also because many messages use figurative or metaphorical expressions. As observed by \citet{muti-etal-2024-pejorativity-disambiguating}, misogyny in Italian is highly metaphorical, especially with references to animals. 
In Italian, not identifying metaphors could also be attributable to translation errors since metaphors are cultural-dependent. 
This highlights the complexity of cross-lingual implicit HS detection, as also pointed out by \citet{korre-etal-2024-challenges-creating}, since the translation of a term often does not carry the same implications as in the source language. 

\paragraph{Wrong Translations.}
This is a category of errors that applies only to Italian.
It comprises errors due to wrong translations of messages or to a lack of understanding of non-standard language, such as dialects and jargon expressions. 


\paragraph{Opposite Intention.} These errors could be considered an instance of LLM hallucinations~\cite{maynez-etal-2020-faithfulness}.
In these cases, the models completely misinterpret the message's content, resulting in generated implied assumptions that tend to support the message. 
These errors occur in both languages, with a slight preference for Italian.

\paragraph{Denial of Misogyny.} 
This class of errors indicates a lack of connection between the generated implied assumptions and the answer of the model. 
In other words, the generated text is logical and correct as it clearly identifies the misogynous nature of the message.
However, the model classifies the texts as non-misogynous.

\paragraph{Lack of Reference}
This kind 
of error has been observed only in Italian - mostly because very few English data contains direct reference to individuals. 
The errors in these cases are due to the model failing to identify the mentioned individuals (and their associated characteristics) that are used to trigger the misogynous content.
In the example we report in Table~\ref{tab:examples}, the model does not recognize (Moana) Pozzi as a famous porn actress, thus resulting in a warrant that fails to capture the insulting nature of the message (i.e., the target being promiscuous).  

\paragraph{Wrong Inference}
This is the largest class of errors in both languages. 
We observe that wrong inferences are mainly driven by spurious correlations and the activation of implicit knowledge. 






\subsection{\fr{How Reasoning Impact Classification}}

We further examine how reconstructed implicit assumptions relate to classification predictions for misogyny detection.

We inspect 50 instances \am{for each language} that are correctly classified by our \am{best} baseline (few-shot \texttt{Llama3-8B}) and that are subsequently misclassified when the model is asked to reason about them. 
We observe the tendency to reduce a gender-related problem to the whole category of human beings, minimizing the misogynistic nature of the statement.
For instance, in the Italian text \textit{``Oggi sei felice anoressica bugiarda''} [Today you are happy anorexic liar], the model responds that \textit{``Anorexia is a negative condition. 3. Lying is a negative behavior. No, the text is not misogynist. Anorexia is a condition that affects both men and women.''}. Although being true that it is a condition that affects both genres, in this case, it targets a woman which is clear in Italian by the gendered \textit{``anoressica''}, instead of \textit{``anoressico''} which would be used for men.
\am{Adding the statement that it affects both genres is detrimental for the classification.}

\section{Conclusion}
We proposed the task of implicit misogyny detection under an Argumentative Reasoning perspective, since to understand implicit statements, one needs to reconstruct the missing link (the warrant) between the claim and the assumption. 
Our work highlights the complexity of such a task, which paves the way for hate speech detection as a proxy task to probe the reasoning abilities of LLMs.
Our prompt-based experiments show
\am{that LLMs fail 68\% and 50\% of the time in generating implied assumptions in Italian and English respectively.}
The poor relationship between wrongly generated explanations and correctly predicted classes shows
LLMs' over-reliance on their implicit knowledge and spurious correlations rather than reasoning skills.
Our results are consistent with \citet{zhu-etal-2023-learn}: prompting strategies that rely on implicit knowledge in LLMs often generate an incorrect classification when the generated knowledge (implied assumptions/warrants) is wrong, due to lack of references, reasoning skills, or understanding of non-standard language. 
Indeed, verifying the validity of the generated text before injecting it in the prompt in a human-in-the-loop approach would be a next step to undertake.
\am{To conclude, our findings show that \textit{i)} the performance of the classification task cannot be used as a proxy to guarantee the correctness of the implied assumption/warrant; \textit{ii}) LLMs do not have the necessary reasoning abilities in order to understand highly implicit misogynistic statements. Therefore, models for hate-related natural language inference tasks should be improved.}
\am{One possible approach would be }to inject external knowledge in the misogynous texts, in order to fill the gaps related to their lack of implicit knowledge. For instance, had the model known that Moana Pozzi was a porn actress, it would have probably inferred that when a person is compared to her, it is a derogatory way to address that woman.

\section*{Limitations}
A limitation of our work is the integration of all generated knowledge (implied assumptions/warrants) and we do not evaluate them before using them to inform the classification task. This should be overcome with a human-in-the-loop approach that allows for the verification of the knowledge extracted by LLMs. We did not try to inject only the knowledge that led to a correct classification because of the low correlation between the generated implied statement and the class.
Another limitation is that for what concerns Llama, many examples in English trigger the safeguard, therefore the scores for Llama might not be realistic.

\section{Ethical Considerations}

Improving LLMs abilities to understand the implied meaning of messages with sensitive content is a case of potential risks related to dual use. Although our work has focused on assessing LLMs abilities in generating implied assumptions/warrants, we see the benefits and the detrimental effects. On the one hand, improving LLMs abilities to understand the implied meaning of sensitive message can further be used to improve the generation of counter-speech and the development of assistive tools for experts in this area. At the same time, the process can be inverted: malevolent agents can feed models with implied assumptions and generate hateful messages. We are aware of this issue, and we think our work offers the community an opportunity to understand limitations of LLMs that have a not minor societal impact. In addition to this, our work indicates the need to adopt different safeguard methods that are able to capture the core meaning of a message and grounded in different cultures.





\bibliography{paper,anthology}

\clearpage
\appendix

\section{ImplicIT-Mis Sources}
\label{sec:app_it_data}

\setcounter{table}{0}
\renewcommand{\tablename}{Table} 
\renewcommand{\thetable}{\Alph{table}} 

Table~\ref{tab:my_label} shows statistics on the number of Facebook comments associated to each newspaper or Facebook community.

\begin{table}[!h]
\centering
\setlength{\tabcolsep}{1.8pt}
\begin{tabular}{lr}
\toprule
\bf Source & \bf Messages \\ \midrule
\bf National News   \\
\,\,\,La Repubblica   & 411 \\
\,\,\,Il Messaggero   & 378 \\
\,\,\,La Stampa       &  76\\
\,\,\,TgCom24         & 20    \\
\,\,\,Libero          & 1 \\ 

\bf Local news  \\
\,\,\,AnconaToday     & 20 \\
\,\,\,BolognaToday    & 9  \\
\,\,\,Corriere Adriatico &  2 \\
\,\,\,Palermolive.it  & 5 \\

\bf Online news     \\
\,\,\,Donna Fanpage   & 37  \\ 
\,\,\,Fanpage         & 33 \\
\,\,\,Huffington Post & 6 \\
\,\,\,TPI             & 4 \\
\,\,\,Il Post         & 1 \\
\,\,\,Leggo           & 1 \\

\bf FB Community    \\
\,\,\,Caffeina  Festival & 65 \\
\,\,\,\makecell[l]{Non sono bello \\ ma spaccio} & 15\\
\,\,\,La matita scarlatta & 9 \\
\,\,\,Pastorizia never dies & 9 \\ 
\,\,\,Stefano Valdegamberi & 6 \\
\,\,\,I love Patriarcato 2 & 4 \\ 
\,\,\,La società femminista & 4 \\
\,\,\,\makecell[l]{L'uomo che \\ bestemmiava ai cavalli}& 3  \\
\bottomrule
\end{tabular}

    \caption{List of sources - newspapers and Facebook pages - with total amount of extracted instance for the creation of the ImplicIT-Mis dataset}
    \label{tab:my_label}
\end{table}

\section{ImplicIT-Mis Annotation Guidelines for Implied Assumptions}
\label{sec:guidelines_it}

Annotators were given the following guidelines for the identification of the implied statements in the ImplicIT-Mis dataset:

\textit{Read the following comments extracted from Facebook as a reaction to a news article or a post. The comments have been judged as being sexist or misogynistic by an expert. Your task is to provide an explanation for why the comment can be offensive towards. The explanation must correspond to the implied assumption in the comment. To help you in this task, imagine that your explanations should complete the sentence: ``This comment is sexist because it implies that \dots ''.}
\textit{Follow the examples.
Le dovrebbero dare 2000 volts implica che la donna in questione dovrebbe essere uccisa.
Non dovremmo abbassare i nostri standard solo per assumere più donne implica che le donne sono meno qualificate degli uomini.
Quella mangia kinder bueno non meritava di vincere X-Factor implica che la donna in questione viene offesa in quanto grassa.
Bau bau implica che la donna in questione viene chiamata cagna.}

Translated version:
\textit{She should be given 2000 volts implied that the targeted woman should be killed.
We shouldn't lower our standards just to hire more women implies that women are less qualified than men.
That Kinder Bueno eater did not deserve to win X-Factor implies that the targeted woman is called 'fat'.
Wof wof implies that the targeted woman is called 'cagna', translated as female dog, which is an epithet for b**ch.}



\clearpage

\section{English Prompts}
\label{sec:prompts_en}

\setcounter{table}{2}
\renewcommand{\tablename}{Table} 
\renewcommand{\thetable}{\Alph{table}} 

\begin{table}[!ht]
    \centering
    \footnotesize
    \setlength{\tabcolsep}{1pt}
    \begin{tabular}{l|l}
    \toprule
        \bf Prompt & \bf Prompt Details \\
        \midrule
        Zero-shot & Is this text misogynous? Answer 1 for yes and 0 for no. \\
        \midrule        
        Few-shot & \makecell[l]{Is the text misogynous? Answer only with YES or NO. Answer like in the examples. \\ Text: we shouldn't lower our standards just to hire more women. \\ Answer: YES \\ Text: But since only women get pregnant, they get free birth control pills. \\ Answer: NO \\  Text: I can't stand women talking about football. \\ Answer: YES \\ Text: Lmao I'm down with women replacing God. The future is female \\ Answer: NO \\ Text: She must be on her period. \\ Answer: YES  \\ Text: Dude, girls want sex as much as guys do. They have triggers, so do we. \\ Answer: NO \\ Text: \{text\} \\ Answer: }\\ \midrule
        \makecell[l]{Implied Assumption \\ Zero-shot} & \makecell[l]{What are the implied assumptions of the text {text}? \\ Based on the implied assumptions, is the text misogynous? \\ Answer only with 1 for yes and 0 for no. \\ Answer between the tags. Implied assumption: <...> Answer: <...> } \\ \midrule
        \makecell[l]{Implied Assumption \\ Few-shot} & \makecell[l]{What are the implied assumptions of the text \{text\}? \\ Based on the implied assumptions, is the text misogynous? \\ Answer only with YES or NO. Answer like in the examples. \\ Text: we shouldn't lower our standards just to hire more women. \\ Implied assumption: women are less qualified than men. \\ Answer: yes. \\ Text: Women shouldn't talk about football. \\ Implied assumption: women cannot understand football, it's a male thing. \\ Answer: YES. \\ Text: She must be on her period. \\ Implied Assumption: women are irrational or overly emotional during their periods. \\ Answer: YES  \\ Text:  \\ Implied assumption: \\ Answer: } \\ \midrule
        \makecell[l]{Toulmin CoT \\ Zero-Shot} & \makecell[l]{What is the claim and the implied warrant of the text \{text\}? Explain step by step.  \\ Based on the implied warrant, is the text misogynous?  Answer only with 1 for yes and 0 for no.  \\ Warrant:  \\ Misogyny: } \\ \midrule
        \makecell[l]{Toulmin CoT \\ Few-Shot} & \makecell[l]{What are the claim and the implied warrant in the input text? Explain step by step. \\ Based on the implied warrant, is the text misogynous? Answer only with YES or NO. Follow the example.  \\  Text: we shouldn't lower our standards just to hire more women \\ Claim: hiring more women would lower our standards \\ Warrant: women are less qualified than men \\ Answer: YES \\ Text: I can't stand women talking about football \\ Claim: women should not talk about football \\  Warrant: football it's for men \\ Answer: YES \\ Text: She must be on her period. \\ Claim: She must be on her period  \\ Warrant: women are irrational or overly emotional during their periods \\ Answer: YES \\  Text: \{text\} \\  Claim: \\  Warrant: \\ Answer: } \\
        
        \bottomrule
    \end{tabular}
    \caption{English promts used in our experiments.}
    \label{tab:prompt}
\end{table}

\clearpage

\section{Italian Prompts}
\label{sec:prompts_it}

\setcounter{table}{3}
\renewcommand{\tablename}{Table} 
\renewcommand{\thetable}{\Alph{table}} 

\begin{table}[!ht]
    \centering
    \footnotesize
    \setlength{\tabcolsep}{1pt}
    \begin{tabular}{l|l}
    \toprule
        \bf Prompt & \bf Prompt Details \\
        \midrule
        Zero-shot & Questo testo è misogino? Rispondi solo con 1 (sì) o 0 (no) \\
        \midrule        
        Few-shot & \makecell[l]{Questo testo è misogino? Rispondi solo con 1 (sì) o 0 (no) come nell'esempio. \\ Testo: quella mangia kinder bueno non meritava di vincere X-Factor. \\  Misogino: 1. \\ Testo: Lei è acida perché non ha figli penso che darebbe fastidio a qualsiasi donna. Che schifo  \\ Risposta: 0  \\ Testo: le donne che parlano di calcio non si possono sentire. \\ Risposta: 1 \\ Testo: Comunque le pringles più buone sono quelle alla panna acida e cipolla \\ Risposta: 0  \\ Testo: Dopo aver visto (per sbaglio) 5 minuti di Uomini e Donne ho deciso che odierò le donne. \\ Risposta: 1 \\ Testo: pur abitando in culo alla balena, sono andata ad un bel po’ di concerti e sono ricordi preziosi\\ che conservo con affetto \\  Risposta:0 \\ Testo: \{text\} \\ Risposta: }\\ \midrule
        \makecell[l]{Implied Assumption \\ Zero-shot} & \makecell[l]{Quali sono gli assunti impliciti del testo \{testo\}? \\ Sulla base degli assunti impliciti, il testo è misogino? \\ Rispondere solo con SÌ o NO. } \\ \midrule
        \makecell[l]{Implied Assumption \\ Few-shot} & \makecell[l]{Quali sono gli assunti impliciti del testo \{testo\}? \\ Sulla base dei presupposti impliciti, il testo è misogino? \\ Rispondere solo con SÌ o NO. Seguire l'esempio. \\ Testo: non dovremmo abbassare i nostri standard solo per assumere più donne. \\ Presupposto: le donne sono meno qualificate degli uomini. \\ Risposta: SÌ. \\ Testo: le donne che parlano di calcio non si possono sentire. \\ Presupposto: le donne non capiscono niente di calcio, è una cosa da maschi. \\ Risposta: SÌ. \\ Testo: Dopo aver visto (per sbaglio) 5 minuti di Uomini e Donne ho deciso che odierò le donne. \\ Presupposto: Le donne che vanno alla trasmissione Uomini e Donne sono stupide. \\ Risposta: SÌ. \\ Testo: \{testo\} \\ Presupposto:  \\ Risposta: \\}  \\ \midrule
        \makecell[l]{Toulmin CoT \\ Zero-Shot} & \makecell[l]{Quali sono il claim e il warrant implicito del testo? Spiegalo passo dopo passo. \\  In base al warrant implicito, il testo è misogino? Rispondi solo con 1 per il sì e 0 per il no. \\ Warrant: \\  Misoginia: } \\ \midrule
        \makecell[l]{Toulmin CoT \\ Few-Shot} & \makecell[l]{Quali sono il claim e il warrant implicito nel testo? Spiegalo passo per passo. \\ In base al warrant implicito, il testo è misogino? Rispondere solo con SÌ o NO. Segui l'esempio. \\ Testo: non dovremmo abbassare i nostri standard solo per assumere più donne. \\ Affermazione:  assumere più donne abbasserebbe i nostri standard \\ Warrant: le donne sono meno  qualificate degli uomini \\ Risposta: SÌ \\  Testo: Non sopporto che le donne parlino di calcio \\ Affermazione: le donne non dovrebbero parlare di calcio \\ Warrant: il calcio è per gli uomini \\ Risposta: SÌ \\ Testo: Deve avere il ciclo. Affermazione: deve avere le mestruazioni \\ Warrant: le donne sono irrazionali o eccessivamente emotive durante il ciclo mestruale \\ Risposta: SÌ \\ Testo: \{testo\} \\ Affermazione: \\ Warrant: \\ Risposta:} \\
         \bottomrule
    \end{tabular}
    \caption{Italian prompts used in our experiments.}
    \label{tab:prompt}
\end{table}

\clearpage

\section{Additional Classification Results}

\setcounter{table}{4}
\renewcommand{\tablename}{Table} 
\renewcommand{\thetable}{\Alph{table}} 

Table~\ref{tab:appendix:results_overview} reports classification results when considering the refusal to answer due to model safeguard trigger to hateful content as misogynous.
In particular, \texttt{Llama3-8B} is the only affected model in our experiments.

\begin{table*}[!t]
    \centering
    \small
    \begin{tabular}{ll|cc}
    \toprule
     \bf Exp. Setting & \bf Model & \bf ImplicIT-Mis & \bf SBIC+ \\ \midrule
    \multirow{2}{*}{fine-tuning} & \texttt{bert-hateXplain} & -- & 0.342 \\
    & \texttt{ALBERTo} & 0.380 & -- \\ \midrule \midrule
    \multirow{2}{*}{zero-shot} & \texttt{Llama3-8B} & 0.588 & 0.609 \\
    & \texttt{Mistral-7B-v02} & 0.050 & 0.319 \\ \midrule
    \multirow{2}{*}{few-shot} & \texttt{Llama3-8B} & \bf 0.738 & \underline{0.827} \\ 
    & \texttt{Mistral-7B-v02} & 0.259 & 0.416 \\ \midrule \midrule

    \multirow{2}{*}{\makecell[l]{zero-shot w.\\ implied assumption}} & \texttt{Llama3-8B} & 0.542 & \underline{0.891} \\
    & \texttt{Mistral-7B-v02} & 0.050 & 0.259  \\ \midrule
    \multirow{2}{*}{\makecell[l]{few-shot w.\\ implied assumption}} & \texttt{Llama3-8B} & 0.480  & \bf \underline{0.914} \\
    & \texttt{Mistral-7B-v02} & 0.461 & 0.685 \\ \midrule \midrule

    \multirow{2}{*}{\makecell[l]{zero-shot \\ Toulmin   warrant}} & \texttt{Llama3-8B} & 0.557 & \underline{0.643} \\
    & \texttt{Mistral-7B-v02} & 0.346 & 0.374 \\ \midrule
    \multirow{2}{*}{\makecell[l]{few-shot \\ Toulmin   warrant}} & \texttt{Llama3-8B} & 0.725 & \underline{0.841} \\
    & \texttt{Mistral-7B-v02} & 0.556 & 0.604 \\ 
    \bottomrule
    \end{tabular}
    \caption{Overview of the results of the experiments on ImplicIT and SBIC+. Best results are in bold, while performance differences with respect to~\ref{tab:results_overview} are underlined.
    Answer considered valid with implied assumption/Toulmin's warrant only if the model generates the implied assumptions/warrants.
    }
    \label{tab:appendix:results_overview}
\end{table*}


\setcounter{table}{5}
\renewcommand{\tablename}{Table} 
\renewcommand{\thetable}{\Alph{table}}

\end{document}